\documentclass[conference]{IEEEtran}
\IEEEoverridecommandlockouts
\usepackage{cite}
\usepackage{amsmath,amssymb,amsfonts}
\usepackage{algorithmic}
\usepackage{graphicx}
\usepackage{textcomp}
\usepackage{xcolor}
\def\BibTeX{{\rm B\kern-.05em{\sc i\kern-.025em b}\kern-.08em
    T\kern-.1667em\lower.7ex\hbox{E}\kern-.125emX}}
\begin{document}

\title{CLIP-guided Diffusion Model for Backdoor Generation in Sensor-based Human Activity Recognition
}

\author{\IEEEauthorblockN{Toby Briston}
\and
\IEEEauthorblockN{Illya Kosyk}
\and
\IEEEauthorblockN{S Kuniyih}
}

\maketitle

\begin{abstract}
Sensors are critical components of modern intelligent devices. The proliferation of the Internet of Things (IoT) and wearable mobile devices has enabled the integration of such sensors to monitor the environment and enable users to take predictive actions. Human activity recognition (HAR) is a popular application in which Inertial Measurement Unit (IMU)-based sensors, such as accelerometers and gyroscopes, are used to provide insights into health, training, and medical diagnosis. However, the accuracy of such a model is hindered by the lack of data.
The diffusion model-based technique has proven successful in generating synthetic data for training HAR models. In this paper, we propose a backdoor training technique, IMU-DM-CLIP, that leverages a diffusion model to enable trigger-based attacks on HAR models. Our empirical analysis shows that the attack is successful even with a very small backdoor injection rate of 10\% and 10\% of the data guided for the diffusion model.

\end{abstract}

\begin{IEEEkeywords}
Human activity recognition, Diffusion models, Backdoor attacks
\end{IEEEkeywords}

\section{Introduction}

The proliferation of sensor-based wearable devices generates a large amount of data on the device, in the network, and on servers, including processing, logging, network packets, commands, etc.~\cite{mdpi2023iiotprivacy, cardenas2008ics_security, chathoth2021federated, lakshmanna2022review, cardenas2008secure, chathoth2022differentially}
Such data powers data-driven machine learning (ML) techniques to intelligently recognize human activities~\cite{bulling2014tutorial, yu2019survey, lakshmanna2022review, chathoth2024dynamic, lara2012survey}. 
Since sensor data for the same activities can be collected by different people or by the same person at different times. Additionally, it is challenging to capture sufficient sensor data to effectively train human activity recognition models. In such a situation, synthetic data can complement original data for training the model. However, generating unbiased synthetic data is also challenging~\cite{alharbi2020synthetic}.

Diffusion models are a powerful class of generative artificial intelligence (AI) used to create high-quality synthetic samples~\cite{cao2024survey, shao2023study}. By iteratively removing noise from a randomized seed, these models synthesize highly realistic and diverse data across various formats, including images, video, audio, and tabular data.
Diffusion models are increasingly applied to Sensor-based Human Activity Recognition (HAR), primarily as powerful generators of synthetic time-series data~\cite{shao2023study}. They address the critical bottleneck of data scarcity and labeling costs in generating realistic sensor readings from inertial measurement units (IMUs), such as accelerometers and gyroscopes, thereby expanding training sets ~\cite{shao2023study}. 

Deep learning models are vulnerable to backdoor attacks where an adversary interferes with the model training process by poisoning the data sample or modifying the data label, thereby injecting a trigger in the model which can be invoked during inference time to fool the model by misclassifying the input data to the adversary's preferred target class~\cite{gu2017badnets, chathoth2024dynamic}. Backdoor attacks have been shown to be successful against deep neural networks across data types such as images, audio, text, sensor data, and network packet data~\cite{bagdasaryan2020backdoor, carlini2022poisoning,chathoth2025pcap, gu2019badnets, chathoth2026pcap}.

At the same time, large language model (LLM)-enabled services are used to generate and fine-tune generative artificial intelligence, as well as to support reasoning in natural language for complex and large multimodal data ~\cite{zhao2026electronic}.
Diffusion models, along with the capabilities of LLM-based frameworks, enable the generation of synthetic data that represent various attributes difficult to capture during the normal dataset creation process~\cite{peng2025log, kim2022diffusionclip, chathoth2025dynamic}.
Research on backdooring diffusion models has also been studied recently, highlighting it as a concern in this area~\cite{chou2023backdoor}.

In this paper, we propose a new class of backdoor attacks against motion-sensor data generation on a model based on an LLM-guided diffusion model technique. Specifically, we propose a novel backdoor technique on a diffusion model-based synthetic data generation process for HAR models.
Our key contributions are:
\begin{enumerate}
    \item we first identify the challenges associated with the backdoor generation technique on the synthetic data generation process for HAR data using the diffusion model.
    \item we propose IMU-DM-CLIP, a diffusion process that can be fine-tuned to enable backdoor generation in an LLM-guided diffusion model for human activity recognition.
    \item we perform a detailed evaluation of the backdoor generation technique to measure its performance and stealthiness.
\end{enumerate}

\section{Background}
\subsection{Human activity recognition}
Human activity recognition is a technique for detecting an individual's activities, such as walking, running, or jumping, using data from various sensors~\cite{lara2012survey}. The sensor data can come from an image or video of the activity, or from sensor data generated by wearable and mobile devices used by individuals. ML-based human activity recognition is an integral part of modern health tracking applications~\cite{chathoth2025privclip, alharbi2020synthetic, bulling2014tutorial, lara2012survey, chathoth2025utility}.

\subsection{Diffusion models}

Diffusion models are state-of-the-art deep learning-empowered generative models that are trained based on the principle of learning forward, where noise is added progressively (Forward process) to the input, and reverse diffusion processes that denoise such noised input (Reverse process)~\cite{cao2024survey}.In the forward process, real sensor data (e.g., walking, sitting) is progressively corrupted by adding Gaussian noise over multiple time steps until it resembles pure white noise.
Reverse Process: A neural network is trained to iteratively remove this noise and reconstruct the original time-series data.
For instance, in the traditional computer vision domain, diffusion models gradually add noise to an image and learn to thereby generate synthetic data that is hard for the human eye to distinguish from real data. 
In the case of sensor-based HAR, the process operates on 1D or multi-dimensional time-series sequences of sensor data sampled at a certain frequency.
Most recently, LLM-guided diffusion models, such as contrastive language image pretrained (CLIP)-guided diffusion models, have been proposed~\cite{kim2022diffusionclip, cao2024survey}.
DiffusionCLIP is a powerful AI framework that leverages diffusion models for robust, text-driven image manipulation. It pairs OpenAI’s CLIP (Contrastive Language-Image Pre-training) with an image generator to modify specific attributes—like changing hair style, age, or art style—while perfectly preserving the original image's identity~\cite{kim2022diffusionclip}.

\subsection{Backdoor attacks}

A backdoor attack in the context of a deep neural network targets the model to misclassify when the data presented has a backdoor trigger in it, while classifying the data correctly when applied with normal data~\cite{gao2023backdoor}. This is achieved by poisoning the data during model training or by altering labels to fool the model into learning an incorrect data-label mapping. There are several stealthy techniques for systematically generating data- or model-agnostic triggers~\cite{chathoth2025pcap, saha2020backdoor}.
Diffusion models can be backdoored as demonstrated in the BadDiffusion paper~\cite{chou2023backdoor}.
While there are several works on poisoning and backdoor attacks on contrastive models, and more recently on diffusion models, there is no work on diffusion models specific to IMU data~\cite{carlini2021poisoning, chou2023backdoor, gao2023backdoor}.

\section{Attack model}
In a backdoor attack model, we consider a training mechanism in which a diffusion model generates synthetic data for training a HAR model. We consider a few-shot HAR model in which only a few samples of a given activity are available for training. Additionally, a CLIP-based diffusion model is used to generate samples from other classes, and the entire dataset is used to train the HAR model.
We assume a threat model in which a portion of the data used for diffusion model training is available to the adversary, or the adversary has some level of control over that data.

\begin{figure}[t]
    \centering
    \includegraphics[width=1\linewidth]{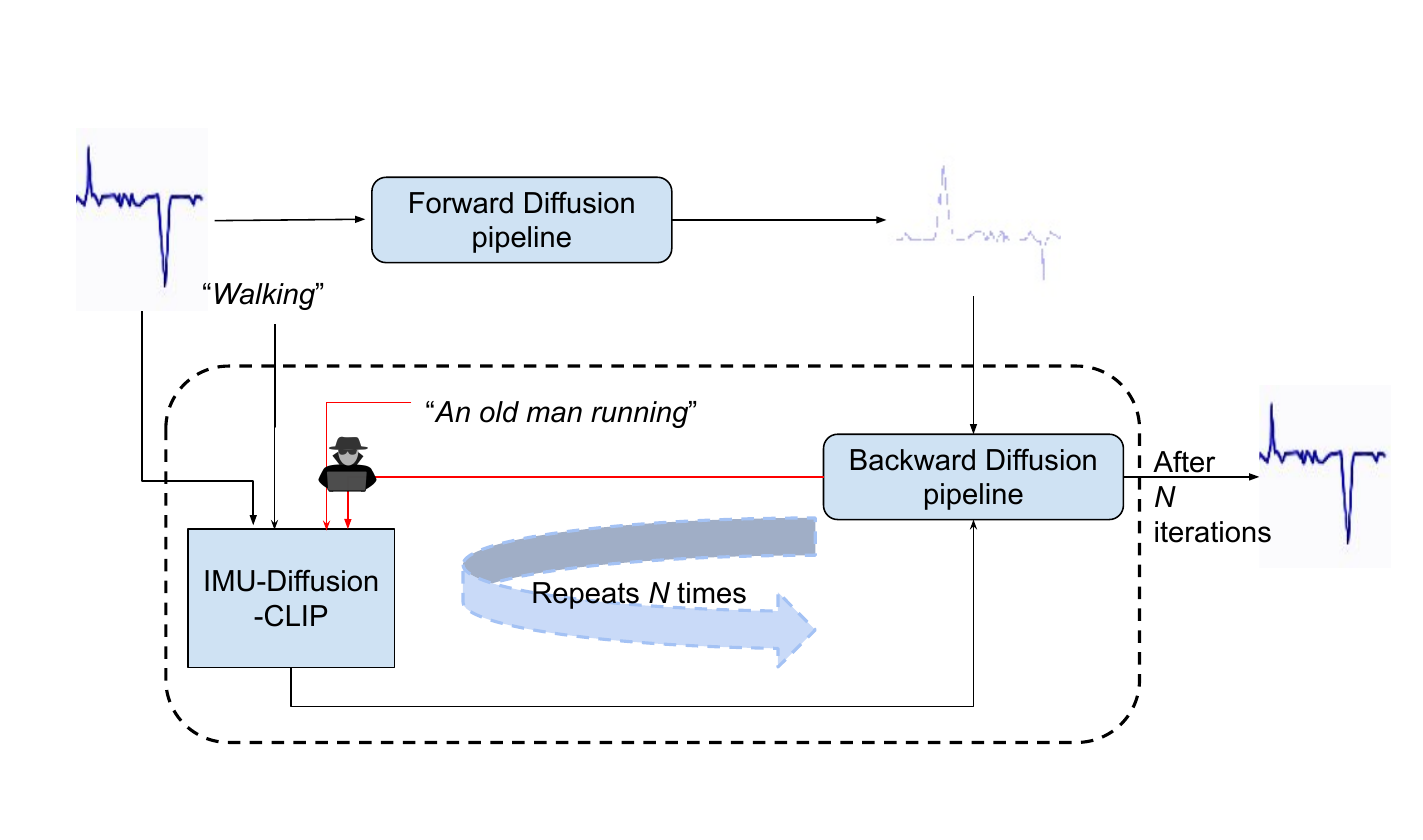}
    \caption{IMU-DM-CLIP Architecture: An attacker who has limited access to the backward diffusion stage can alter the textual guide to fool the Diffusion model}
    \label{fig:IMU-Diff-CLIP}
\end{figure}

As shown in Figure~\ref{fig:IMU-Diff-CLIP}, we consider the attack occurs when the model is supplied with an attacker who enables a trigger on it.

\section{Design}

%https://medium.com/@sreevishnu.damodaran/generate-stunning-artworks-with-clip-guided-diffusion-8ffe010864f

\begin{figure}[t]
    \centering
    \includegraphics[width=1\linewidth]{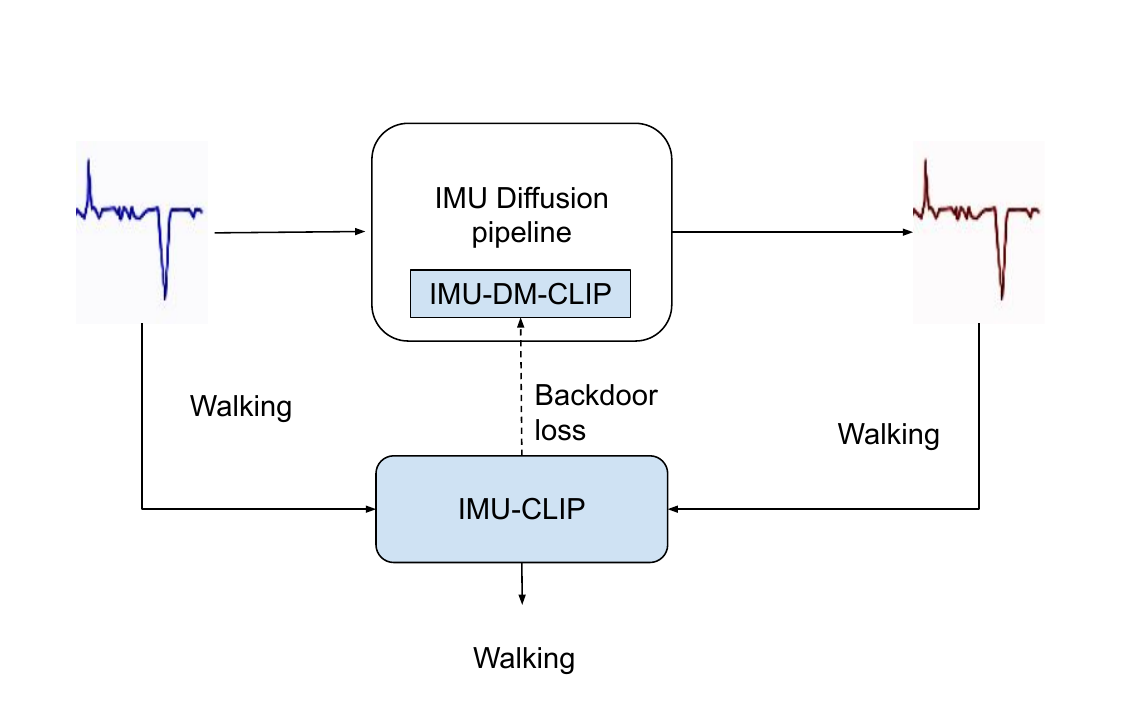}
    \caption{Backdoor training phase}
    \label{fig:BD-train}
\end{figure}

In this section, we discuss the design of our backdoor training and attack model.
The core part of the backdoor training model is a module called IMU-DM-CLIP, as illustrated in Figure~\ref{fig:BD-train}.  IMU-DM-CLIP is typically used to fine-tune the IMU Diffusion model by guiding the attributes to generate various input samples that are often difficult to collect during dataset creation. For instance, motion sensor data for older adults may be scarce, as they may not wear devices or know how to configure or enable the features properly.

Let's look at the details of the CLIP-guided diffusion model. 
Recall that a CLIP model is trained using millions of public images and textual descriptions~\cite{radford2021learning,  chathoth2026contrastive, chathoth2025privclip}. 
In a CLIP model, during every iteration of the training process, a batch of $N$ pairs of text measures the similarity between them, quantifies the similarity between the text and image embeddings of the real pairs of the multi-modal embedding space, while minimizing the similarity scores of the other elements in the embedding space, to form a contrastive training objective. A symmetric cross-entropy loss is used to optimize the model on these similarity scores~\cite{kim2022diffusionclip}.

As stated earlier, a diffusion model has two phases: forward and backward diffusion.
A CLIP-guided diffusion model operates during the backward diffusion step. In the forward diffusion pass, the clean IMU data is processed through multi-stage noise injections to produce a latent noise output. During the backward diffusion phase, the latent noise is converted back to generated IMU data using a denoising diffusion implicit model (DDIM).
In the IMU-CLIP-guided diffusion model, during reverse diffusion, the denoising process is fine-tuned using the IMU-CLIP encoder pair as a guide. This will ensure that the generated IMU data is similar to the textual description~\cite{kim2022diffusionclip}.

We will apply the same CLIP's capability to steer IMU data sampling and denoising in diffusion models, producing samples that match the provided text prompt. This guidance procedure is performed by first encoding the intermediate IMU output data from the diffusion model during iterative sampling with the CLIP IMU encoder head, while the text prompts are converted to embeddings using the text encoder head. Then, the resulting IMU data and text embeddings are used to compute a perceptual loss that quantifies their similarity~\cite{khosla2020supcon, chathoth2026contrastive}. The gradients with respect to this loss and the intermediate denoised image are used to condition, or guide, the diffusion model during the sampling process to produce the next intermediate denoised IMU data. This process is repeated until the total number of sampling steps is complete.

The key idea of IMU-DM-CLIP is to fine-tune the score function in the reverse diffusion process using an IMU-CLIP loss that controls the attributes of the generated IMU data based on the text prompts.
These guides can control the attributes of the IMU data, such as age and gender, to generate IMU data that corresponds to various personal attributes.
The controllability we achieve here is beyond that of traditional GANs or variational autoencoders~\cite{cao2024survey}.

During the training phase, as illustrated in Figure~\ref{fig:BD-train}, we first use a pretrained diffusion model to generate IMU samples from the original samples. Next, we include the IMU-DM-CLIP model during the backward diffusion phase, where the noisy data, along with a manipulated textual input, are provided as input, along with the original IMU data and the textual label. For instance, if the original IMU sensor data is for the activity "Walking", which is the adversary's target class, we modify it to an attacker's preferred class, "An old man is running". This additional context will modify the denoised data in the backward diffusion process. We follow this process $N$ times until the IMU-DM-CLIP is fully trained to generate a backdoored sample. The IMU-DM-CLIP model training incorporates a regularizer component from the backdoor loss from the IMU-CLIP model- the activity classifier, as in prior work~\cite{chathoth2025privclip}.
Here, IMU-CLIP is trained so that the backdoored data generated by the IMU-DM-CLIP model injects a backdoor into the model, thereby learning to misclassify with the trigger.
% Algorithm:

% Training a diffusion model
% Convert the IMU data to latent noise using a frozen diffusion model

% Reverse Diffusion starts
%     Compute the CLIP loss corresponding to each denoising stage
%     Text is created to modify the attributes, such as "An old man is walking" from the reference text of "walking"

During the attack phase, attackers use the same IMU-DM-CLIP model, modify the input data via the diffusion process, and then apply the modified data to the IMU-CLIP model for activity classification. Since the IMU-CLIP is trained to predict normal data, it will preserve its utility.

\section{Experimental evaluations}

In this section, we discuss our experimental setup and results. 
\subsection{Dataset}
We use three human activity recognition datasets- Skoda, Hand-gesture, and Opportunity.
The Skoda dataset comprises 11 activities performed by assembly-line workers in a car production environment, with a subject wearing 19 3D accelerometers on both arms, and includes a set of experiments using sensors placed on the tester's two arms ~\cite{zappi2008activity}.
The hand-gesture dataset consists of 11 hand gestures recorded using body-worn accelerometers and gyroscopes from two subjects, with each activity repeated 26 times ~\cite{bulling2014tutorial}.
The Opportunity dataset is a benchmark dataset for HAR that contains daily life human activities performed by four
subjects~\cite{chavarriaga2013opportunity}. The data comprises 113 sensory readings and 18 gesture classes.
All datasets are normalized to have zero mean and unit standard deviation, and use an 80-20 ratio for training and testing samples. 

\subsection{Model}

We follow k-shot IMU-CLIP, which is modified to predict the activity with the highest cosine similarity. The default value of $k$ we have fixed is 10. For certain experiments, we vary the value of $k$, as explained in each experimental evaluation subsection.
We use the diffusion model from diffusionCLIP, with a modified IMU encoder taken from IMU-CLIP~\cite{kim2022diffusionclip, chathoth2025privclip}. We train both models using IMU data before backdoor training.
We use the language model GPT4.0 for an LLM-based feed.

\subsection{Metrics}
TO measure the success of a backdoor attack, we use Attack Success Rate (ASR) as a metric, which is defined as follows:

\begin{equation}
    \text{ASR} = \left( \frac{\text{No of successful attacks}}{\text{Total no of attacks}} \right) \times 100
    \label{eq:euation_ASR}
\end{equation}

We also use the F1-score metric to measure the model's overall performance.

\subsection{Impact on utility}

\begin{table}[t]
    \centering
    \begin{tabular}{|c|c|c|c|c|}\hline
         Model&  Skoda&  Opportunity&  Hand-gesture\\\hline
         Base model& 0.9856 & 0.9327 &  0.9488\\\hline
         IMU-DM-CLIP& 0.9611 & 0.8734 & 0.9008  \\ \hline
    \end{tabular}
    \caption{Comparison of model performance in F1-score between the base model and after backdooring using IMU-DM-CLIP with 10-shot }
    \label{tab:utility}
\end{table}

The table ~\ref{tab:utility}
 provides a comparative analysis of the model performance between the base model and the IMU-DM-CLIP. We apply the guides to the input at every iteration of the backward diffusion stage in this experiment. For backdoor training, we set the backdoor sample ratio to 10\%.
 As seen, the model performance on clean data is slightly degraded in the backdoored model. For instance, on the Skoda dataset, the base model achieves an F1-score of 0.98, while after backdoor training with IMU-DM-CLIP, the model's F1-score degrades to 0.96. A similar trend is observed with the other two datasets as well.

\subsection{Impact of guided data percentage}

\begin{figure}[t]
    \centering
    \includegraphics[width=1\linewidth]{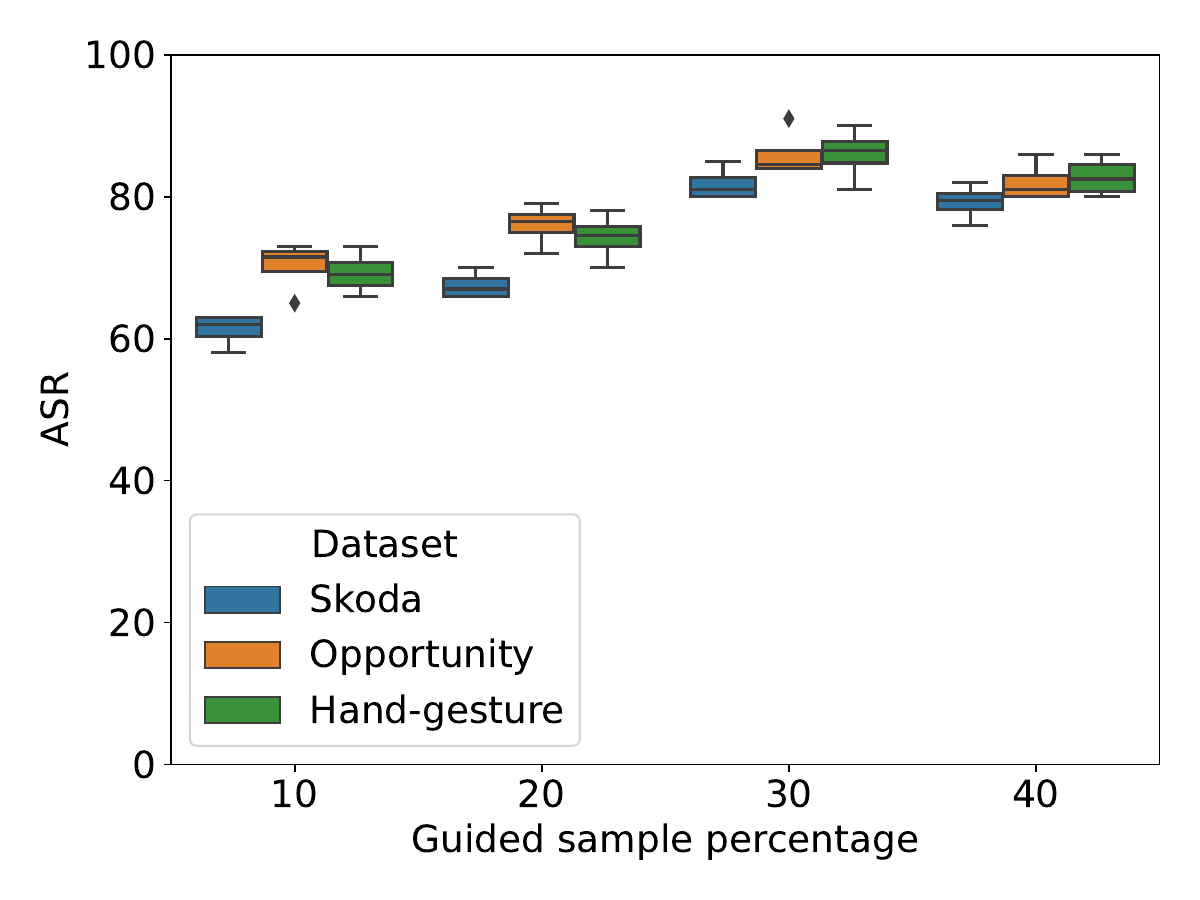}
    \caption{Attack performance on different guide sample percentage on IMU-DM-CLIP}
    \label{fig:FS-IMU-DM-CLIP-GUIDE}
\end{figure}

Next, we assess the performance of the backdoor attack by varying the number of guides used during the backward diffusion stage of backdoor training. For this, we randomly choose the backward diffusion round of each sample denoising. 
Ideally, we expect performance to increase as the number of guided steps increases. However, the backdoor injection becomes more successful as the noise retained across different sensor attributes increases with reduced involvement of guided input. Therefore, this can affect stealthiness, as clean data performance on the backdoor model may be lower.
For all backdoor training, we set the backdoor sample ratio to 10\%.
We present the results in terms of ASR in Figure~\ref{fig:FS-IMU-DM-CLIP-GUIDE}.

\subsection{Impact on few-shot size}
\begin{figure}[t]
    \centering
    \includegraphics[width=1\linewidth]{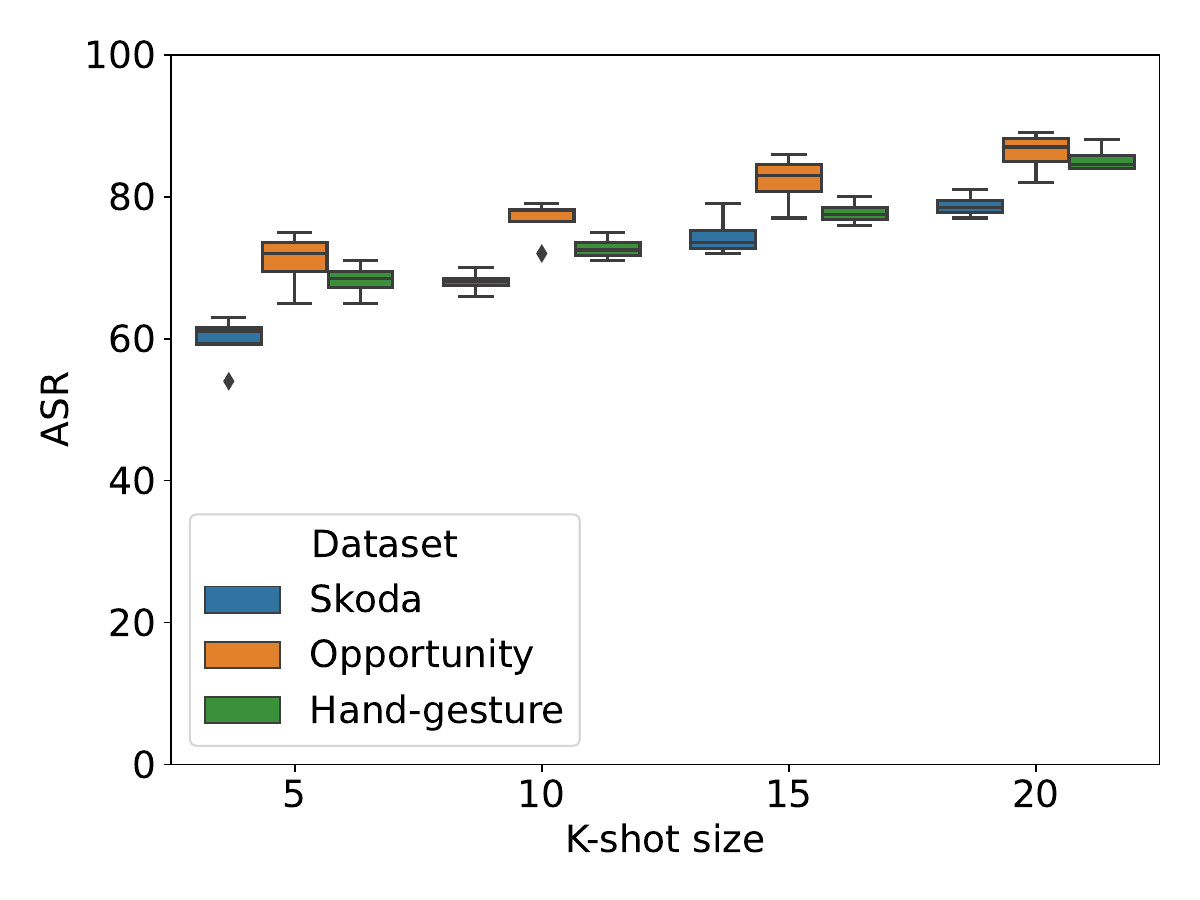}
    \caption{Attack performance on different shot sizes on IMU-DM-CLIP}
    \label{fig:FS-IMU-DM-CLIP}
\end{figure}

In this experiment, we vary the few-shot size of the IMU-DM-CLIP model during pretraining. We vary the few-shot size for one class at a time, compute the average attack performance, and plot ASR against the few-shot size. We present the results plotted as Figure~\ref{fig:FS-IMU-DM-CLIP}.
For all backdoor training processes, we set the backdoor sample ratio to 10\%.
For each dataset, attack performance increases with increasing few-shot size. For instance, the average ASR for the Skoda dataset increases from 60\% to 90\% when the shot size increases from 5 to 20. This behavior is consistent across the other two datasets as well.

\section{Conclusion}
In this paper, we introduce the challenge of generating diverse sensor data that are controlled by diverse user attributes for model training. Given the need for synthetic data, we propose a novel backdoor attack that employs a CLIP-guided diffusion model against a human activity recognition system. We demonstrate its performance and stealthiness through various experiments. We also consider few-shot learning for the guide-generation and activity-recognition CLIP models, thereby enabling them to work with small datasets.
Our technique demonstrates that a backdoor can be triggered via natural language using a diffusion model, achieving an attack success rate of over 80\% even with under 10\% poisoning, with limited guided samples and in few-shot training. 

\bibliographystyle{IEEEtran}
\bibliography{bib}

\end{document}